\documentclass[runningheads]{llncs}
\usepackage[T1]{fontenc}
\usepackage{graphicx}
\usepackage{amsmath}
\usepackage{tikz}
\usetikzlibrary{shapes.geometric, arrows.meta, positioning, fit, backgrounds, shadows, calc}

\usepackage{url}
\begin{document}
%
\title{CAF-Gen: A Multi-Agent System for \\Enriching Argumentation Structures}

%
%
\author{Jakub Bąba\orcidID{0009-0009-7000-4887} \and
Jarosław A. Chudziak\orcidID{0000-0003-4534-8652}}
%
\authorrunning{J. Bąba and J. A. Chudziak}
%
\institute{Faculty of Electronics and Information Technology,\\ Warsaw University of Technology, Poland \\
\email{\{jakub.baba.stud,jaroslaw.chudziak\}@pw.edu.pl}}
%
\maketitle              
\begin{abstract}
Formalizing complex reasoning from natural text is one of the central challenges in computational linguistics. It requires systems to understand not just keywords but also the context and complex reasoning embedded in a text. Current Argument Mining (AM) techniques identify basic claims and premises, yet they often struggle to capture the richer structural information required by advanced schemas such as the Carneades Argumentation Framework (CAF), which incorporates features such as premise types, proof standards, and argument schemes. We address this limitation by introducing CAF-Gen, an automated multi-agent framework designed to enrich shallow argument structures into CAF-compliant argument models. By employing an iterative Creator-Reviewer pipeline, a creator agent’s output is validated by a critical agent to ensure structural integrity. This multi-agent collaboration is crucial for mitigating the structural instability typical of single-pass generative models. Our experiments demonstrate that the iterative feedback loop improves the quality of the resulting data and achieves strong alignment with the original annotations, while producing structurally richer models. Our findings show that the multi-agent system can overcome the limitations of single-pass generation, providing a robust methodology for the automated modeling of formal argumentation.

\end{abstract}

\section{Introduction}
Argument Mining~\cite{lawrence-reed-2019-argument} investigates the automatic extraction and structuring of arguments from natural language texts. While this field advanced rapidly in identifying basic argumentation graphs, typically consisting of claims, premises and simple attack or support relations, current AM techniques predominantly focus on these shallow models, leaving the extraction of more nuanced information, included in advanced formal argumentation frameworks like ASPIC+~\cite{modgil2014aspic+} or the Carneades Argumentation Framework (CAF)~\cite{gordon2006carneades} underexplored. The Carneades Argumentation Framework offers a sophisticated model that incorporates features such as distinct premise types, varied proof standards, and explicit argument schemes. The extraction of the arguments using frameworks such as CAF is typically beyond the scope of standard AM tasks, primarily due to a lack of established, annotated corpora necessary for developing and validating such systems for these complex structures. Consequently, the practical application of these extensive features and frameworks within computational intelligence remains limited.

Large Language Models (LLMs) have shown improved performance in these tasks, promising opportunities to address this challenge. Their strengths in processing and generating natural language, paired with their ability to learn from contexts, have led to an increasing adoption of solutions incorporating LLMs in demanding NLP tasks~\cite{ruiz2024nlas}. However, single-pass generation by LLMs often struggles with the strict constraints required by formal logic, leading to hallucinations. To overcome this issue, the concept of Multi-Agent Systems (MAS) has emerged, where the interaction between agents can simulate collective intelligence to refine complex outputs. By decomposing a task into specialized, well-organized subtasks, MAS allow each agent to focus on a narrower objective, enabling more targeted quality control.

In this paper, we introduce CAF-Gen, a multi-agent LLM-driven framework designed to automatically enrich basic argument annotations into rigorous, CAF-compliant models. Our primary contribution is the application and validation of an iterative Creator-Reviewer pipeline specifically adapted for the domain of formal argumentation in order to ensure both the structural integrity and semantic richness of the generated models. Based on this approach, we demonstrate the system's efficacy by retrieving detailed features from the UKP Argument Annotated Essays~\cite{stab2014annotating} corpus and enriching them into CAF-compliant models.

\section{Background and Related Work}

\begin{figure}
\includegraphics[width=\textwidth]{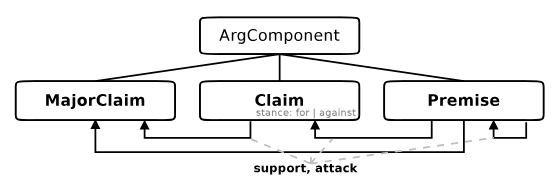}
\caption{Annotation scheme introduced by Stab and Gurevych~\cite{stab2014annotating}.}
\label{fig:influential_schema}
\end{figure}

Argument Mining (AM) has emerged as an important research area in Natural Language Processing (NLP) and computational linguistics. It is focused on the automated identification and modeling of the argumentative structures within text and extensively surveyed in ~\cite{lawrence-reed-2019-argument}~\cite{lippi2016argumentation}~\cite{vecchi-etal-2021-towards}. Foundational work in AM was focused on providing annotation schemes, allowing for structuring and indexing retrieved annotations. One influential scheme, introduced by Stab and Guryevich\cite{stab2014annotating}, distinguishes claims, premises, and their support or attack relations (see Figure \ref{fig:influential_schema}). One of the annotated corpora, building on that scheme, was a corpus of persuasive essays called UKP Argument Annotated Essays~\cite{stab-gurevych-2017-parsing}. This corpus and other notable corpora, like AIFdb~\cite{lawrence2012aifdb}, have been crucial to developing and benchmarking argument mining systems. While these corpora are invaluable for benchmarking core AM tasks, the diversity and quality of AM corpora are continuously further enhanced. The areas of improvement include multimodal and multilingual corpora and research~\cite{mancini-etal-2024-mamkit}~\cite{toledo-ronen-etal-2020-multilingual}~\cite{sermpezis2025agoraspeechmultiannotatedcomprehensivedataset}.

For truly enabling automated argument formalization in domains like law and science, the underlying index must capture a deeper level of reasoning.  While frameworks like ASPIC+~\cite{modgil2014aspic+} offer rigorous logical structuring based on inference rules, we selected the Carneades Argumentation Framework (CAF)~\cite{gordon2007carneades} as our target because of its alignment with natural language semantics. CAF explicitly structures arguments around argument schemes (e.g. "Argument from Expert Opinion" and "Argument from Example", defined in argument scheme taxonomies such as Walton's~\cite{walton2013argumentation}). Furthermore, CAF defines features such as statement types (ordinary premises, assumptions, and exceptions) and proof standards that guide a statement's acceptability based on the evidential burden provided in a given context. This richness enables the modeling of more sophisticated reasoning compared to simple schemas. However, this complexity introduces a significant challenge: mapping raw text to these structured features remains a problem which limits the applications of CAF in computational systems.

Large Language Models (LLMs), such as GPT-4~\cite{achiam2023gpt} have opened new possibilities for NLP and AM tasks~\cite{chen-etal-2024-exploring-potential}, offering a path to bridge this schema gap. LLMs have demonstrated previously unseen capabilities of understanding and generation of natural language and context and have been shown to vastly outperform previous models in AM tasks, whether through special prompting techniques or fine-tuning~\cite{gorur2024largelanguagemodelsperform}~\cite{cabessa-etal-2025-argument}~\cite{guida2025llmsargumentminingdetection}. Among these techniques, one work presented a modular framework based on structured prompting for legal rule application~\cite{sadowski2025explainableruleapplicationstructured}. The proposed method decomposed reasoning tasks into manageable steps and enabled formal validation of the LLM-generated outputs against logical constraints. Their ability to deduce implicit information and generate structured output~\cite{ruiz2024nlas} makes them promising candidates for generating the complex features required by frameworks like CAF. Building on these capabilities, the concept of multi-agent systems (MAS) has emerged~\cite{tran2025multiagentcollaborationmechanismssurvey}~\cite{zhao2025siriusselfimprovingmultiagent}. This agentic approach,
where each agent has a structured role, has been presented in various works in
knowledge management and NLP, including~\cite{maslowski2026heterogeneousdebateengineidentitygrounded}~\cite{10.1007/978-981-96-6008-7_21}. One of the foundational works in this area is the CRITIC framework, which allows LLMs to validate and progressively improve their responses using tools~\cite{gou2024criticlargelanguagemodels}. Another recent work introduced ACC-Collab, a framework based on an Actor-Critic approach that produces a two-agent team that focuses on collaboration~\cite{estornell2025acccollabactorcriticapproachmultiagent}.

\section{Problem and Approach}
To address the challenge of automated formalization, we designed CAF-Gen, a multi-agent framework that transforms input text annotated with basic argument structures into a rich, structurally valid structure compliant with the Carneades schema. The process is built on a collaborative, iterative pipeline between two LLM agents: a Creator and a Reviewer. This architecture ensures high projection to the original structure while reliably enriching the model with the complex features required for formal reasoning.




\subsection{Task Formulation}
 
We formalize CAF enrichment as a structure-preserving graph
transformation that maps a shallow argument graph onto a
Carneades-compliant one, enriching every node and edge with the
attributes required for formal reasoning while remaining grounded in the
input structure.
 
The \textbf{source graph} is defined as follows:
\begin{equation}
  G_{src} = \langle V_{src},\, E_{src} \rangle
\end{equation}
It consists of components $V_{src}$ and directed relations $E_{src}$:
\begin{align}
  \tau_{src} &: V_{src} \to
    \{\mathsf{MajorClaim}, \mathsf{Claim}, \mathsf{Premise}\}, \\
  E_{src} &\subseteq V_{src} \times V_{src} \times
    \{\mathsf{Support}, \mathsf{Attack}\}.
\end{align}
This representation encodes argumentative role and polarity, but none of
the logical attributes required for computational acceptability under CAF.
 
The \textbf{target graph} is defined as follows:
\begin{equation}
  G_{CAF} = \langle S,\, A \rangle
\end{equation}
It consists of statements $S$ and arguments $A$. Each statement $s \in S$
is enriched with a statement type $\sigma(s) \in T_{stmt}$ and a proof
standard $\pi(s) \in P_{std}$, where $T_{stmt}$ and $P_{std}$ are the CAF
statement-type and proof-standard taxonomies~\cite{gordon2006carneades}.
Each argument $a \in A$ is a tuple
$a = \langle \mathit{prem}(a), c(a), \mathit{pol}(a), w(a) \rangle$ with
premises $\mathit{prem}(a) \subseteq S$, conclusion $c(a) \in S$,
polarity $\mathit{pol}(a) \in \{\mathsf{Pro}, \mathsf{Con}\}$, and an
argument scheme $w(a)$ from Walton's taxonomy\cite{walton2013argumentation}.
 
The objective is to construct an enrichment map
\begin{equation}
  \Phi : G_{src} \to G_{CAF}
\end{equation}
subject to two grounding constraints. \emph{Component grounding} assigns
every source component a statement via a map $\mu : V_{src} \to S$ (major
claims becoming argument conclusions), fixing $\sigma$ and $\pi$; $\mu$
need not be injective, as redundant components may be consolidated.
\emph{Relation grounding} maps every relation $(u, v, r) \in E_{src}$ to
an argument with conclusion $\mu(v)$ whose polarity preserves $r$
($\mathsf{Support} \mapsto \mathsf{Pro}$,
$\mathsf{Attack} \mapsto \mathsf{Con}$). The central challenge is that
$\sigma$, $\pi$, and $w$ are absent from $G_{src}$ and must be inferred
from the text, while $\Phi$ stays anchored to the original structure
rather than introducing ungrounded nodes or edges.

\subsection{The Enrichment Workflow}

\begin{figure}
    \centering
    \includegraphics[width=\textwidth]{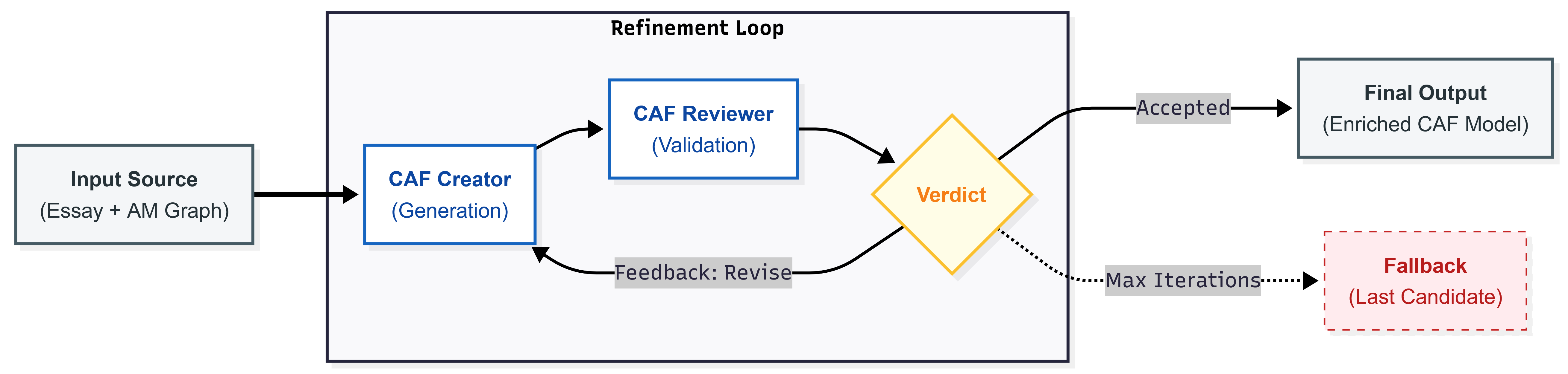}
    \caption{The CAF-Gen system overview.}
    \label{fig:system_overview}
\end{figure}

CAF-Gen is a framework designed to automate this enrichment process. It is grounded on two LLM agents: the CAF Creator and the CAF Reviewer. These agents work iteratively to produce a structured index entry out of each unit from the input corpus. The overview of this system is presented in Figure \ref{fig:system_overview}.

The CAF Creator and CAF Reviewer operate together in an iterative refinement loop to increase the quality of the generated CAF models. After each creation of the model, it is passed to the Reviewer for evaluation. If the model is accepted, the process is finalized. However, when issues appear, warnings, suggestions, and refinements are provided and resubmitted to the Creator along with the model and input data. The whole cycle of generation, review, and revision is continued until the Reviewer accepts the model or the system exceeds predefined number of iterations. The whole process aims to progressively improve the accuracy of the output model and detect errors.

The dataset used in the experiments of our methodology is the full \textbf{UKP Argument Annotated Essays v2}~\cite{stab-gurevych-2017-parsing} corpus. This publicly available corpus contains 402 persuasive essays, annotated coherently with earlier introduced~\cite{stab2014annotating} components: Major Claims, Claims, and Premises, along with the argumentative relations between them, categorized as Support or Attack. The widespread adoption of the subset in the Argument Mining studies makes it a valuable and trustworthy benchmark for our research, and the persuasive nature of the essays provides room for identifying and enriching more nuanced components.

\subsection{Agents: Iterative Generation and Quality Assurance}
The CAF Creator is the generative agent in the pipeline. Its responsibility is to take a unit from the corpus and construct a candidate entry according to the CAF schema. This process involves several subtasks:
\begin{enumerate}
\item Component Mapping - Identifying and mapping claims and premises to CAF statements, with Major Claims mapped to conclusions.
\item Type Classification - Assigning a statement type to each statement, such as ordinary premise, presumption, or exception, based on its role within the argument.
\item Proof Standard Assignment - Selecting a proof standard from a predefined set~\cite{inbook} (e.g., Scintilla of Evidence, Preponderance of the Evidence, Beyond Reasonable Doubt) based on the textual context and statement type.
\item Argument Mapping - Identifying and mapping support/attack relations to CAF arguments.
\item Scheme Identification - Determining the most plausible argument scheme from Walton's taxonomy~\cite{walton2013argumentation} (e.g., Argument from Consequences, Argument from Expert Opinion) for each argument.
\end{enumerate}

The CAF Reviewer acts as a critic in the pipeline. It evaluates the structure generated by the CAF Creator and ensures its quality before it is accepted as the final index. The key checks (with the thorough sub-checks) could be defined as follows:
\begin{enumerate}
\item Integrity \& Validity - validating if whole structure is structurally valid, e.g., all argument identifiers are unique and all features are provided
\item Textual Coherence \& Accuracy - based on whole context, checking if the features are appropriately assigned
\item Completeness - validating if all the components are included in the structure and if there are any unexplained changes compared to the input structure
\end{enumerate}

\subsection{Illustrative Example}
To demonstrate the semantic gap between basic argument mining and formal modeling, consider a system attempting to evaluate the statement: \textit{"Does the attack on the mandatory uniform policy use the 'Argument from Negative Consequences' scheme?"}

\begin{figure}
\centerline{\includegraphics[width=0.8\textwidth]{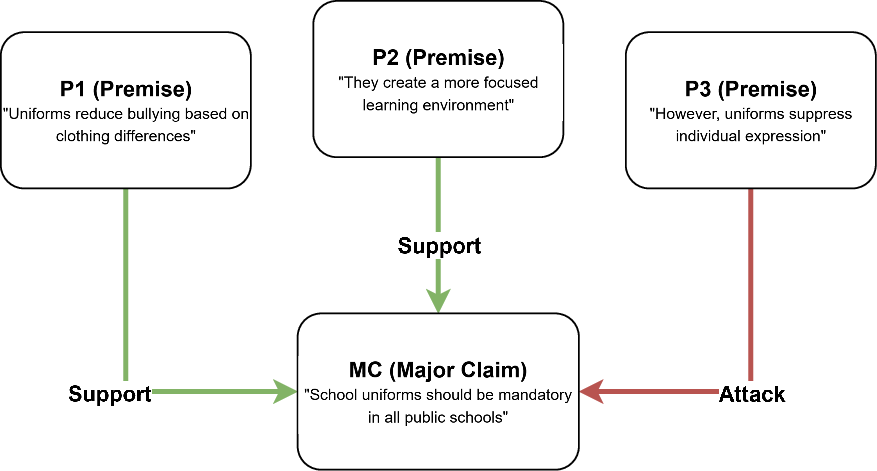}}
\caption{Visualization of the general model.}
\label{fig:basic}
\end{figure}

A system using our basic input model, as shown in Figure \ref{fig:basic}, identifies that P3 attacks the main claim. However, it cannot understand the type of reasoning used and treats it as a generic negative edge. This representation, while valuable, lacks the depth provided by CAF-based models that allow answering this question precisely.

\begin{figure}
\centerline{\includegraphics[width=0.8\textwidth]{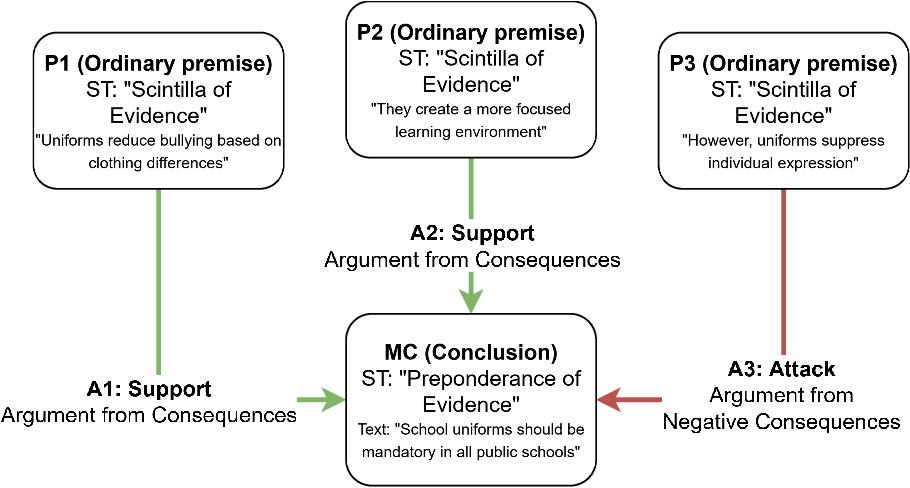}}
\caption{Visualization of the enriched CAF model.}
\label{fig:enriched}
\end{figure}

The CAF-Gen system takes this basic structure as an input and transforms it into an enriched model. As shown in Figure \ref{fig:enriched}, all of the statements have assigned features: statement types and proof standards. All relations are remapped to include an argument type.

The enriched structure, while faithful to the original one, provides richer features, enabling a more rigorous and formal analysis. The system can now retrieve the proper argument looking for an Attack relation and specific argument scheme. The core challenge for our work is to automate this enrichment process accurately and consistently across a range of argument structures.

\section{Experiments and Results}

We conducted a series of experiments to evaluate the CAF-Gen framework against two primary research questions: (Q1) How effective is the iterative refinement process? (Q2) How faithfully does the enriched model preserve the argumentative structure of the source annotations? This section details our setup and presents the quantitative results for each question.

\subsection{Experimental Setup}

Our experiments were conducted on the full \textbf{UKP Argument Annotated Essays v2} corpus, a dataset containing 6089 statements and 3832 relations. Prior to processing by the CAF-Gen, the corpus annotations were converted into a structured JSON format, which is easily parsed by LLM agents. The final output of the CAF-Gen was also generated in JSON.

For the purpose of evaluation (Q2), we designed a deterministic, rule-based projection to map our enriched CAF index back to the simpler source schema for direct comparison. This projection works as follows: Firstly, all CAF statements are projected back to the generalized Component. Secondly, CAF arguments with type Pro are projected to the Support relations, while the Con ones are projected to the Attack relations. Finally, in the original corpus, text is also normalized, and Major Claims, Claims, and Premises are flattened to a Component type. During the projection, all CAF-specific enrichments are discarded during the projection, as they could not be compared with the input format. This projected output serves as the basis for calculating the accordance metrics for Q2. These metrics quantify structural preservation rather than the correctness of the newly assigned CAF features, as no large-scale ground truth currently exists. High scores in these metrics indicate that the enrichment process respect the original argumentation graph without discarding core components, providing an important baseline for validity. 

The CAF Creator and CAF Reviewer agents within the CAF-Gen system were instances of Large Language Models, specifically Google's \textbf{Gemini 2.5 Pro} model, with the temperature set to \textbf{0.1} to promote deterministic outputs, as the task requires strict adherence to the predefined CAF schema. The multi-agent pipeline was implemented using LangGraph and orchestrated through the Gemini API. The CAF-Gen architecture is designed to be model-agnostic, with agents defined solely by their roles and prompt structures, allowing any sufficiently capable LLM to serve as the underlying engine. We selected Gemini 2.5 Pro as a representative of the current state-of-the-art in general purpose large language models, capable of complex instruction-following and structured output generation. The primary goal of this study was to validate the performance of the CAF-Gen architecture itself, rather than to perform a comparative benchmark of different models. The Creator was guided by a detailed prompt instructing it to adhere strictly to the predefined CAF schema, with all the required fields and taxonomies. The Reviewer prompts followed a checklist format, and it returned a summary with the descriptions for each point, each with a marker being one of: \textit{Error, Warning, Suggestion, Info, Accept}. Besides that, the Reviewer decided if the model was accepted, basing on a general rule that Errors and Warnings were subjects to refinement, while summaries containing only other types of messages were leading to acceptance. The iterative refinement loop between CAF Creator and CAF Reviewer was configured to not extend \textbf{five} iterations per argumentative essay. If the Reviewer did accept the solution within these iterations, the model was accepted and proceeded to further validation.

To quantitatively answer our research questions, we defined the following metrics:
\begin{enumerate}
    \item For \textbf{Q1} (Refinement Effectiveness): We measured the impact of the refinement process using:
    \begin{enumerate}
        \item \textbf{First-Pass Acceptance Rate}: The percentage of models that were accepted on the first attempt, without needing any further revisions.
        \item \textbf{Reviewer acceptance rate}: The percentage of models that were accepted within the five-iteration limit.
        \item \textbf{Average iterations}: The mean number of Creator-Reviewer cycles per accepted essay.
        \item \textbf{Distribution of Reviewer Errors and Warnings}: The categorical breakdown of issue flagged by the Reviewer.
    \end{enumerate}
    \item For \textbf{Q2} (Index Fidelity): We measured the accordance of the projected output with the source corpus using standard metrics: \textbf{Precision (P)}, which assesses the accuracy of the generated items (the fraction of generated statements/relations that correctly match the ground truth); \textbf{Recall (R)}, which measures coverage (the fraction of original statements/relations preserved by the workflow); and the \textbf{F1-score (F1)}, the harmonic mean of P and R. We calculated these for:
    \begin{enumerate}
        \item \textbf{Component Identification}: Comparing the set of projected statements against the original claims and premises.
        \item \textbf{Relation Identification}: Comparing the set of Support/Attack relations against the original relations.
    \end{enumerate}
\end{enumerate}

\subsection{Q1: Refinement Process Results}

To answer Q1, we analyzed the interactions between CAF Creator and CAF Reviewer agents. Table \ref{tab:reviewer_feedback} summarizes key performance statistics of the refinement loop and the categorical distribution of different kinds of issues flagged by the Reviewer.

\begin{table}
\caption{CAF Reviewer Feedback and Generation Process Statistics}
\centering
\begin{tabular}{|l|c|}
\hline
\textbf{Statistic / Issue Category} & \textbf{Value / Frequency} \\
\hline
Total Essays Analyzed & 402 \\
First-Pass Acceptance Rate & 34.6\% \\
Final Acceptance Rate (Overall) & 91.3\% \\
Average Iterations per Essay & 2.35 \\
\hline
\textbf{Most Common Reviewer-Flagged Issues:}$^{\mathrm{a}}$&  \\ \hline
\quad Inappropriate Argument Scheme Selection & 38.9\% \\ 
\quad Incorrect Proof Standard Assignment & 9.4\% \\ 
\quad Structural Inconsistency / Invalid Link & 44.7\% \\ 
\quad Schema Violation (Missing/Incorrect Fields) & 2.0\% \\ 
\quad Other Semantic Coherence Issues & 0.6\% \\
\hline
\multicolumn{2}{p{0.95\columnwidth}}{$^{\mathrm{a}}$Percentages for issues are relative to the total number of 'error', 'warning', and 'suggestion' items.}
\end{tabular}
\label{tab:reviewer_feedback}
\end{table}

The first-pass acceptance rate was 34.6\%, which rose to 91.3\% after refinement. The average number of iterations was 2.35. The most common issues were Structural Inconsistency (44.7\%) and Inappropriate Argument Scheme Selection (38.9\%).

\subsection{Q2: Accordance Results of the Projected Structure}

To answer Q2, we evaluated the fidelity of final, accepted index entries by projecting them back to the original UKP annotated corpus. Table \ref{tab:full_identification} presents the Precision, Recall, and F1-score achieved during the comparison of the output model after projection against the original corpus. In the summary, we only included essays that were finally accepted by a Reviewer within a defined number of iterations. 

\begin{table}
\caption{Identification Performance Against Source Corpus}
\centering
\begin{tabular}{|c|c|c|c|}
\hline
\textbf{Evaluation Metric}&\multicolumn{3}{|c|}{\textbf{Component Identification Scores}} \\
\cline{2-4} 
\textbf{} & \textbf{\textit{Precision}}& \textbf{\textit{Recall}}& \textbf{\textit{F1-score}} \\
\hline
Projected CAF Statements & 99.8\% & 98.8\% & 99.3\% \\
Projected CAF Relations & 67.1\% & 99.1\% & 80.0\% \\
\hline
\end{tabular}
\label{tab:full_identification}
\end{table}

For \textbf{Component Identification}, the system achieved a Precision of 99.8\%, Recall of 99.8\% and F1-score of 99.3\%. For \textbf{Relation Identification}, the system achieved a Recall of 99.1\%, Precision of 67.1\% and F1-score of 80.0\%. This lower Precision, along with a very high Recall, score suggests that the workflow not only preserves original relations but also introduces new ones. 

\section{Evaluation and Discussion}
After performing all the experiments, we thoroughly validated the generated corpus. In order to precisely evaluate the results, we reviewed and performed an analysis of the results. Besides that, we performed a manual check on a selected subset, verifying the correctness of the data and furthermore answering our questions.

\subsection{Refinement Process Effectiveness}

The results for Q1, which showed a significant increase in acceptance rate from 34.6\% to 91.3\%, quantitatively demonstrate the necessity of the iterative refinement loop. The analysis of this feedback reveals that the iterative refinement has played a crucial role in quality assurance, making the output models more precise and scrupulous. The final acceptance rate confirms that the vast majority of the structures could be defined within 5 iterations. Additionally, the average of 2.35 iterations per unit shows that most issues were resolved within one or two revisions. These results highlight the efficiency of the LLM-to-LLM feedback loop.

Our manual analysis on a subset of 10 corrected essays where we compared the pairs of first iteration models with the final models confirms these findings. In all of the pairs, the quality of the final model was without a doubt higher in comparison to a first, rejected model. While first models were not precise enough or contained other issues, we found that the final models were plausible. Our analysis proved that the Reviewer was successfully pointing out issues including, but not limited to: incorrect argument schemes, statement types, and proof standards; orphaned and circular relations; redundant statements. In most of the cases, these suggestions improved the quality of the output model drastically, and without them, models would lack the necessary level of depth and sophistication.

To further illustrate the Reviewer's capabilities, let us present an instance where the Creator generated an argument with "Argument from Positive Consequences" as the scheme. The Reviewer flagged this part, arguing about the assigned scheme by providing the part of the sentence 'Without following trends, such people would have a hard time...', pointing to the negative consequence "after not following trends", and suggesting "Argument from Negative Consequences". In the next iteration of the Creator, such issue was fixed. As shown in the table, a significant amount of the issues was suggesting this type of changes, and oftentimes the Creator adjusted the scheme to a more appropriate one.

To mitigate the risk of the Reviewer rejecting valid but stylistically different interpretations, we calibrated the sensitivity by categorizing feedback from blocking "Errors" to non-blocking "Suggestions". Despite this, our analysis revealed some limitations. On several occasions, the Reviewer's feedback was debatable or overly strict, potentially leading to the rejection of a plausible model. This indicates that LLM Reviewers, while powerful, might still benefit from an arbitration mechanism for subjective or debatable cases.

\subsection{Accordance with Source Corpus}
The quantitative results indicate both an extremely high level of fidelity in maintaining the original argumentative components and an ability to make intelligent argumentative enrichments. Component identification scores near 99\% demonstrate that the system maintains the original structure, with minor deviations typically arising from minor consolidations and subtle changes. For relation identification, the F1-score of 80\% is a result of a very high recall score (99.1\%) and a significantly lower precision score (67.1\%). This indicates that the system preserves most of the originally annotated relations, while proactively adding new connections that were implicit in the source. Overall, our system shows a strong capacity to use provided annotations as a foundation and still enrich the model. However, it is important to clarify that these metrics quantify structural fidelity to the original annotations. While they do not assure absolute correctness of the new CAF features, they confirm that our enrichment process respects the underlying structures without destructive behavior.

To validate this assumption, we conducted a detailed analysis of 10 essays, comparing the final models generated by CAF-Gen with the input models. Each of the analyzed essays consisted of differences in the projected statements or relations. In 8 of the 10 essays, we rated the differences in the output model generated by the system positively or very positively, as the system added significant nuance or simplified unnecessary information. A clear example of enrichment was in the essay named "Dormitory or apartment?" where in the original text, a statement "both living accommodations have their own benefits" was an isolated claim. The CAF-Gen system inferred a new \textit{argument from balancing} that linked statements "living in dormitory is always helpful to students" and "living off campus prepares students for the real life" as the premises to the above conclusion. The other essay, named "Saving/Spending money" is a great example of simplification. The input structure consisted of the following Major Claims: "saving some part of your earnings is essential" and "there is no other choice than saving money for some time". Our system correctly detected the semantic overlap and left only the first statement, creating a more concise model without any information loss.  

However, the analysis highlighted the areas for improvement. We rated a single essay as slightly negative, as the system proposed redundant arguments, and even though the model was valid, it was also excessive and cluttered. We also found one neutral case, which revealed that there exist different modeling philosophies - our system favored grouping several statements into a single, strong argument against the input's more numerous one-to-one relations. This case showed that some models, while different, might be equally valid.

\section{Future Work}
Presented experiments confirm that an agentic system based on LLMs can reliably construct complex formal argumentation models and bridge the gap between simple argumentative structures and rigorous logical frameworks.

The most important next step would be to leverage the generated models and explore their possibilities by integrating them into an automated reasoning system. This would allow for empirical testing of how the enriched features enable more sophisticated analysis. Furthermore, it would allow to answer a crucial question of real-world utility and correctness of the approach. While we conducted a manual analysis on a sample of essays in this work, we are developing a Human-in-the-Loop framework, which would be a key part of the comparison between a fully automated framework and one including manual verification. 

Another vital research direction would be a comprehensive benchmarking of the framework's engine. While the study focused on the architecture and investigated it using a single state-of-the-art LLM, a comparative analysis across different LLM families could help in understanding the impact of model choice. Future work will benchmark the performance of different models within the pipeline, allowing one to not only compare the quality of the output, but also to investigate different areas, like the analysis of potential biases in the models.

Finally, future work should also focus on the architecture and refinement process itself. For instance, a concept of multi-agent systems could be further investigated. The two-agent pipeline could be expanded to include specialized agents for different tasks. Improvements in the loop could include, but are not limited to: adopting tools for logical reasoning, including logical solvers or theorem provers, training and fine-tuning specialized models for specific sub-tasks, and involving other agent configurations. One of the promising directions could be incorporating external knowledge through Retrieval Augmented Generation (RAG). By equipping the Creator with a retrieval mechanism over a set of scheme definitions and usage examples, the system could ground its classifications, increasing the overall quality. Ultimately, the aim is to progress towards powerful end-to-end systems that could generate thorough and valid formal argument structures from raw text, without the need to use external annotations.

\section{Conclusion}
Automating formalization of natural language arguments requires models that can capture deep reasoning features, yet reliably constructing such models remains challenging for computational intelligence. To address this, we introduced CAF-Gen, a structured agentic LLM framework, and demonstrated its potential in bridging the gap between basic argument annotations and the rich, detailed structures demanded by the Carneades Argumentation Framework.

Our primary contribution is a methodology that addresses the reliability problems of single-pass LLM generation. Our experiments demonstrate that the iterative Creator-Reviewer pipeline is essential, increasing the acceptance rate of generated models from 34.6\% to 91.3\% by addressing and correcting critical semantic and structural errors. Furthermore, we demonstrated that the resulting index is highly faithful to the source text, while also providing valuable, high quality enrichments. We verified the generated annotations and analyzed their accordance with the input components and relations, proving a high quality of the output corpus. Taken together, these findings confirm that a multi-agent system can reliably automate the creation of a complex argument index.

By connecting natural language argumentation with formal reasoning schemas, this work contributes to the development of intelligent systems for domains requiring precision and formal structure, such as legal reasoning or academic debate. Crucially, we reveal the potential of multi-agent LLM systems to ensure output quality - an area previously believed to require human-level judgment. Ultimately, this research marks a significant step toward the broader integration of formal reasoning, potentially redefining how argumentative texts are understood in automated systems.

\subsubsection{\ackname} The work reported in this paper was supported by the Polish National Science Centre under grant 2024/06/Y/HS1/00197.

%
%
\bibliographystyle{splncs04}
\bibliography{bibliography}

@article{lawrence-reed-2019-argument,
    title = "Argument Mining: A Survey",
    author = "Lawrence, John  and
      Reed, Chris",
    journal = "Computational Linguistics",
    volume = "45",
    number = "4",
    month = dec,
    year = "2019",
    address = "Cambridge, MA",
    publisher = "MIT Press",
    url = "https://aclanthology.org/J19-4006/",
    doi = "10.1162/coli_a_00364",
    pages = "765--818"
}

@article{ruiz2024nlas,
  title={Nlas-multi: A multilingual corpus of automatically generated natural language argumentation schemes},
  author={Ruiz-Dolz, Ramon and Taverner, Joaquin and Lawrence, John and Reed, Chris},
  journal={Data in Brief},
  volume={57},
  pages={111087},
  year={2024},
  publisher={Elsevier}
}

@article{stab-gurevych-2017-parsing,
    title = "Parsing Argumentation Structures in Persuasive Essays",
    author = "Stab, Christian  and
      Gurevych, Iryna",
    journal = "Computational Linguistics",
    volume = "43",
    number = "3",
    month = sep,
    year = "2017",
    address = "Cambridge, MA",
    publisher = "MIT Press",
    url = "https://aclanthology.org/J17-3005/",
    doi = "10.1162/COLI_a_00295",
    pages = "619--659"
}

@inbook{inbook,
author = {Gordon, Thomas and Walton, Douglas},
year = {2009},
month = {05},
pages = {239-258},
title = {Proof Burdens and Standards},
isbn = {978-0-387-98196-3},
journal = {Argumentation in Artificial Intelligence},
doi = {10.1007/978-0-387-98197-0_12}
}

@book{walton2013argumentation,
  title={Argumentation schemes for presumptive reasoning},
  author={Walton, Douglas},
  year={2013},
  publisher={Routledge}
}

@inproceedings{gordon2006carneades,
  title={The Carneades argumentation framework--using presumptions and exceptions to model critical questions},
  author={Gordon, Thomas F and Walton, Douglas},
  year={2006}
}

@inproceedings{stab2014annotating,
  title={Annotating argument components and relations in persuasive essays},
  author={Stab, Christian and Gurevych, Iryna},
  booktitle={Proceedings of COLING 2014, the 25th international conference on computational linguistics: Technical papers},
  pages={1501--1510},
  year={2014}
}

@incollection{lawrence2012aifdb,
  title={AIFdb: Infrastructure for the argument web},
  author={Lawrence, John and Bex, Floris and Reed, Chris and Snaith, Mark},
  booktitle={Computational models of argument},
  pages={515--516},
  year={2012},
  publisher={IOS Press}
}

@article{modgil2014aspic+,
  title={The ASPIC+ framework for structured argumentation: a tutorial},
  author={Modgil, Sanjay and Prakken, Henry},
  journal={Argument \& Computation},
  volume={5},
  number={1},
  pages={31--62},
  year={2014},
  publisher={SAGE Publications Sage UK: London, England}
}

@article{gordon2007carneades,
  title={The Carneades model of argument and burden of proof},
  author={Gordon, Thomas F and Prakken, Henry and Walton, Douglas},
  journal={Artificial intelligence},
  volume={171},
  number={10-15},
  pages={875--896},
  year={2007},
  publisher={Elsevier}
}

@misc{maslowski2026heterogeneousdebateengineidentitygrounded,
      title={Heterogeneous Debate Engine: Identity-Grounded Cognitive Architecture for Resilient LLM-Based Ethical Tutoring}, 
      author={Jakub Masłowski and Jarosław A. Chudziak},
      year={2026},
      eprint={2603.27404},
      archivePrefix={arXiv},
      primaryClass={cs.AI},
      url={https://arxiv.org/abs/2603.27404}, 
}

@misc{gorur2024largelanguagemodelsperform,
      title={Can Large Language Models perform Relation-based Argument Mining?}, 
      author={Deniz Gorur and Antonio Rago and Francesca Toni},
      year={2024},
      eprint={2402.11243},
      archivePrefix={arXiv},
      primaryClass={cs.CL},
      url={https://arxiv.org/abs/2402.11243}, 
}

@misc{tran2025multiagentcollaborationmechanismssurvey,
      title={Multi-Agent Collaboration Mechanisms: A Survey of LLMs}, 
      author={Khanh-Tung Tran and Dung Dao and Minh-Duong Nguyen and Quoc-Viet Pham and Barry O'Sullivan and Hoang D. Nguyen},
      year={2025},
      eprint={2501.06322},
      archivePrefix={arXiv},
      primaryClass={cs.AI},
      url={https://arxiv.org/abs/2501.06322}, 
}

@article{achiam2023gpt,
  title={Gpt-4 technical report},
  author={Achiam, Josh and Adler, Steven and Agarwal, Sandhini and Ahmad, Lama and Akkaya, Ilge and Aleman, Florencia Leoni and Almeida, Diogo and Altenschmidt, Janko and Altman, Sam and Anadkat, Shyamal and others},
  journal={arXiv preprint arXiv:2303.08774},
  year={2023}
}

@inproceedings{chen-etal-2024-exploring-potential,
    title = "Exploring the Potential of Large Language Models in Computational Argumentation",
    author = "Chen, Guizhen  and
      Cheng, Liying  and
      Luu, Anh Tuan  and
      Bing, Lidong",
    editor = "Ku, Lun-Wei  and
      Martins, Andre  and
      Srikumar, Vivek",
    booktitle = "Proceedings of the 62nd Annual Meeting of the Association for Computational Linguistics (Volume 1: Long Papers)",
    month = aug,
    year = "2024",
    address = "Bangkok, Thailand",
    publisher = "Association for Computational Linguistics",
    url = "https://aclanthology.org/2024.acl-long.126/",
    doi = "10.18653/v1/2024.acl-long.126",
    pages = "2309--2330"
}

@inproceedings{cabessa-etal-2025-argument,
    title = "Argument Mining with Fine-Tuned Large Language Models",
    author = "Cabessa, J{\'e}r{\'e}mie  and
      Hernault, Hugo  and
      Mushtaq, Umer",
    editor = "Rambow, Owen  and
      Wanner, Leo  and
      Apidianaki, Marianna  and
      Al-Khalifa, Hend  and
      Eugenio, Barbara Di  and
      Schockaert, Steven",
    booktitle = "Proceedings of the 31st International Conference on Computational Linguistics",
    month = jan,
    year = "2025",
    address = "Abu Dhabi, UAE",
    publisher = "Association for Computational Linguistics",
    url = "https://aclanthology.org/2025.coling-main.442/",
    pages = "6624--6635"
}

@article{lippi2016argumentation,
  title={Argumentation mining: State of the art and emerging trends},
  author={Lippi, Marco and Torroni, Paolo},
  journal={ACM Transactions on Internet Technology (TOIT)},
  volume={16},
  number={2},
  pages={1--25},
  year={2016},
  publisher={ACM New York, NY, USA}
}

@inproceedings{vecchi-etal-2021-towards,
    title = "Towards Argument Mining for Social Good: A Survey",
    author = "Vecchi, Eva Maria  and
      Falk, Neele  and
      Jundi, Iman  and
      Lapesa, Gabriella",
    editor = "Zong, Chengqing  and
      Xia, Fei  and
      Li, Wenjie  and
      Navigli, Roberto",
    booktitle = "Proceedings of the 59th Annual Meeting of the Association for Computational Linguistics and the 11th International Joint Conference on Natural Language Processing (Volume 1: Long Papers)",
    month = aug,
    year = "2021",
    address = "Online",
    publisher = "Association for Computational Linguistics",
    url = "https://aclanthology.org/2021.acl-long.107/",
    doi = "10.18653/v1/2021.acl-long.107",
    pages = "1338--1352"
}

@misc{gou2024criticlargelanguagemodels,
      title={CRITIC: Large Language Models Can Self-Correct with Tool-Interactive Critiquing}, 
      author={Zhibin Gou and Zhihong Shao and Yeyun Gong and Yelong Shen and Yujiu Yang and Nan Duan and Weizhu Chen},
      year={2024},
      eprint={2305.11738},
      archivePrefix={arXiv},
      primaryClass={cs.CL},
      url={https://arxiv.org/abs/2305.11738}, 
}

@misc{estornell2025acccollabactorcriticapproachmultiagent,
      title={ACC-Collab: An Actor-Critic Approach to Multi-Agent LLM Collaboration}, 
      author={Andrew Estornell and Jean-Francois Ton and Yuanshun Yao and Yang Liu},
      year={2025},
      eprint={2411.00053},
      archivePrefix={arXiv},
      primaryClass={cs.CL},
      url={https://arxiv.org/abs/2411.00053}, 
}

@inproceedings{10.1007/978-981-96-6008-7_21,
author = {Harbar, Yarolsav and Chudziak, Jaros\l{}aw A.},
title = {Simulating Oxford-Style Debates with LLM-Based Multi-Agent Systems},
year = {2025},
isbn = {978-981-96-6007-0},
publisher = {Springer-Verlag},
address = {Berlin, Heidelberg},
url = {https://doi.org/10.1007/978-981-96-6008-7_21},
doi = {10.1007/978-981-96-6008-7_21},
booktitle = {Intelligent Information and Database Systems: 17th Asian Conference, ACIIDS 2025, Kitakyushu, Japan, April 23-25, 2025, Proceedings, Part I},
pages = {286–300},
numpages = {15},
location = {Kitakyushu, Japan}
}

@inproceedings{mancini-etal-2024-mamkit,
    title = "{MAMK}it: A Comprehensive Multimodal Argument Mining Toolkit",
    author = "Mancini, Eleonora  and
      Ruggeri, Federico  and
      Colamonaco, Stefano  and
      Zecca, Andrea  and
      Marro, Samuele  and
      Torroni, Paolo",
    editor = "Ajjour, Yamen  and
      Bar-Haim, Roy  and
      El Baff, Roxanne  and
      Liu, Zhexiong  and
      Skitalinskaya, Gabriella",
    booktitle = "Proceedings of the 11th Workshop on Argument Mining (ArgMining 2024)",
    month = aug,
    year = "2024",
    address = "Bangkok, Thailand",
    publisher = "Association for Computational Linguistics",
    url = "https://aclanthology.org/2024.argmining-1.7/",
    doi = "10.18653/v1/2024.argmining-1.7",
    pages = "69--82"
}

@inproceedings{toledo-ronen-etal-2020-multilingual,
    title = "Multilingual Argument Mining: Datasets and Analysis",
    author = "Toledo-Ronen, Orith  and
      Orbach, Matan  and
      Bilu, Yonatan  and
      Spector, Artem  and
      Slonim, Noam",
    editor = "Cohn, Trevor  and
      He, Yulan  and
      Liu, Yang",
    booktitle = "Findings of the Association for Computational Linguistics: EMNLP 2020",
    month = nov,
    year = "2020",
    address = "Online",
    publisher = "Association for Computational Linguistics",
    url = "https://aclanthology.org/2020.findings-emnlp.29/",
    doi = "10.18653/v1/2020.findings-emnlp.29",
    pages = "303--317"
}

@misc{sermpezis2025agoraspeechmultiannotatedcomprehensivedataset,
      title={AgoraSpeech: A multi-annotated comprehensive dataset of political discourse through the lens of humans and AI}, 
      author={Pavlos Sermpezis and Stelios Karamanidis and Eva Paraschou and Ilias Dimitriadis and Sofia Yfantidou and Filitsa-Ioanna Kouskouveli and Thanasis Troboukis and Kelly Kiki and Antonis Galanopoulos and Athena Vakali},
      year={2025},
      eprint={2501.06265},
      archivePrefix={arXiv},
      primaryClass={cs.CL},
      url={https://arxiv.org/abs/2501.06265}, 
}

@misc{guida2025llmsargumentminingdetection,
      title={LLMs for Argument Mining: Detection, Extraction, and Relationship Classification of pre-defined Arguments in Online Comments}, 
      author={Matteo Guida and Yulia Otmakhova and Eduard Hovy and Lea Frermann},
      year={2025},
      eprint={2505.22956},
      archivePrefix={arXiv},
      primaryClass={cs.CL},
      url={https://arxiv.org/abs/2505.22956}, 
}

@misc{sadowski2025explainableruleapplicationstructured,
      title={Explainable Rule Application via Structured Prompting: A Neural-Symbolic Approach}, 
      author={Albert Sadowski and Jarosław A. Chudziak},
      year={2025},
      eprint={2506.16335},
      archivePrefix={arXiv},
      primaryClass={cs.AI},
      url={https://arxiv.org/abs/2506.16335}, 
}

@misc{zhao2025siriusselfimprovingmultiagent,
  title={SiriuS: Self-improving Multi-agent Systems via Bootstrapped Reasoning},
  author={Wanjia Zhao and Mert Yuksekgonul and Shirley Wu and James Zou},
  year={2025},
  eprint={2502.04780},
  archivePrefix={arXiv},
  primaryClass={cs.AI},
  url={https://arxiv.org/abs/2502.04780},
}

\end{document}